\pdfoutput=1

\documentclass[11pt]{article}

\usepackage{ACL2023}

\usepackage{times}
\usepackage{latexsym}

\usepackage[T1]{fontenc}

\usepackage[utf8]{inputenc}

\usepackage{microtype}
\usepackage{diagbox}
\usepackage{slashbox}

\usepackage{inconsolata}

\usepackage{amsmath}
\usepackage{amsfonts}
\usepackage{graphics}
\usepackage{subfigure}
\usepackage{graphicx}
\usepackage{multirow}
\usepackage{colortbl}
\usepackage{xcolor}
\newcommand{\model}{\text{DiffuSum}}
\newcommand\Tstrut{\rule{0pt}{2.6ex}}         
\newcommand\Bstrut{\rule[-0.9ex]{0pt}{0pt}}   

\title{DiffuSum: Generation Enhanced Extractive Summarization with Diffusion}

\author{Haopeng Zhang$^{\ast}$, Xiao Liu\thanks{ $\quad$equal contribution} , Jiawei Zhang \\       IFM Lab, Department of Computer Science, University of California, Davis, CA, USA \\\texttt{haopeng,xiao,jiawei@ifmlab.org}}

\begin{document}
\maketitle
\begin{abstract}
Extractive summarization aims to form a summary by directly extracting sentences from the source document. Existing works mostly formulate it as a sequence labeling problem by making individual sentence label predictions. This paper proposes DiffuSum, a novel paradigm for extractive summarization, by directly generating the desired summary sentence representations with diffusion models and extracting sentences based on sentence representation matching. In addition, DiffuSum jointly optimizes a contrastive sentence encoder with a matching loss for sentence representation alignment and a multi-class contrastive loss for representation diversity. Experimental results show that {\model} achieves the new state-of-the-art extractive results on CNN/DailyMail with ROUGE scores of $44.83/22.56/40.56$. Experiments on the other two datasets with different summary lengths also demonstrate the effectiveness of {\model}. The strong performance of our framework shows the great potential of adapting generative models for extractive summarization. To encourage more following work in the future, we have
released our codes at \url{https://github.com/hpzhang94/DiffuSum}
\end{abstract}
\section{Introduction}
Document summarization aims to compress text material while keeping its most salient information. It plays a critical role with the growing amount of publicly available text data. Automatic text summarization approaches can be divided into two streams: abstractive and extractive summarization. Although abstractive methods \cite{nallapati2016abstractive,gupta2019abstractive, bae2019summary,li2020keywords} produce flexible and less redundant summaries, they suffer from problems of generating ungrammatical or even nonfactual contents \cite{kryscinski2019evaluating,zhang2022improving}. In contrast, extractive summarization forms a summary by directly extracting sentences from the source document. Thus, the extracted summaries are grammatically accurate and faithful.

We focus on extractive summarization in this work. Extractive summarization is commonly formulated as a sequence labeling problem, which predicts a $0/1$ label for each sentence, indicating whether the sentence should be included in summary~\cite{nallapati2017summarunner,zhou2018neural,liu2019text}. Compared to individual sentence label prediction in the sequence labeling setting, generative models offer increased flexibility and attend to the entirety of input context.
Recent works have also successfully applied generative models to wide-ranging token-level sequence labeling tasks~\cite{athiwaratkun2020augmented,du2021all,yan2021unified}. Nonetheless, how to apply generative models for sentence-level tasks like extractive summarization has not been explored.

Recently, continuous diffusion models have achieved great success in vision and audio domains~\cite{ho2020denoising,kong2020diffwave,yang_diffusion_2022,rombach2022high,ho2022video}. Researchers have also attempted to apply diffusion models for text generation by converting the discrete token to continuous embeddings and mapping from embedding space to words with a rounding method~\cite{li_diffusion-lm_2022, yuan2022seqdiffuseq,strudel2022self,gong_diffuseq_2022}. However, these approaches are not applicable for sentence-level tasks like summarization: \textbf{(1)} Summarization has a relatively longer input context and larger generation length (around 3-6 sentences), while the above token-level diffusion-LM models are only applicable to short generation tasks like text simplification and question generation. Their performance tends to drop by a large margin when generating longer sequences; \textbf{(2)} The word embeddings generated by these models could be indistinguishable, resulting in ambiguous and hallucinated generation; \textbf{(3)} The rounding step in those existing diffusion models is less efficient and slows down the inference dramatically. 

To address the above issues, we propose a novel extractive summarization paradigm, \textbf{\model}, which generates the desired summary sentence representations with transformer-based diffusion models and extracts summaries based on sentence representation matching. Instead of generating word by word, \model \ directly generates the desired continuous representations for each summary sentence and thus could process much longer text. \model \ is a summary-level framework since the transformer-based diffusion architecture generates all summary sentence representations simultaneously. Moreover, \model \ incorporates a contrastive sentence encoding module with a matching loss for sentence representation alignment and a multi-class contrastive loss~\cite{khosla2020supervised} for representation diversity. \model \ jointly optimizes the \textit{sentence encoding module} and the \textit{diffusion generation module}, and extracts sentences by representation matching without any rounding step. We validate \model \ by extensive experiments on three benchmark datasets and experimental results demonstrate that \model \ achieves a comparable or even better performance than state-of-the-art systems that rely on pre-trained language models. \model \ also shows a strong adaptation ability based on cross-dataset evaluation results. 

We highlight our contributions in this paper as follows:

\textbf{(i)} We propose \model, a novel generation-augmented paradigm for extractive summarization with diffusion models. \model \ directly generates the desired summary sentence representations and then extracts sentences based on representation matching. To the best of our knowledge, this is the first attempt to apply diffusion models for the extractive summarization task.

\textbf{(ii)} We also introduce a contrastive sentence encoding module with a matching loss for representation alignment and a multi-class contrastive loss for representation diversity. 

\textbf{(iii)} We conduct extensive experiments and analysis on three benchmark summarization datasets to validate the effectiveness of \model. \model \ achieves new extractive state-of-art results on CNN/DailyMail dataset with ROUGE scores of $44.83/22.56/40.56$.


\section{Related Work}
\subsection{Extractive Summarization}
Recent advances in deep neural networks have dramatically boosted the progress in extractive summarization systems. Existing extractive summarization systems span an extensive range of approaches. Most works formulate the task as a sequence classification problem and use sequential neural models with different encoders like recurrent neural networks~\cite{cheng2016neural,nallapati2016abstractive} and pre-trained language models~\cite{egonmwan2019transformer,liu2019text,zhang2023extractive}. Another group of work formulates extractive summarization as a node classification problem and applies graph neural networks to model inter-sentence dependencies ~\cite{xu2019discourse,Zhang2020TextGT,wang2020heterogeneous,zhang2022hegel}. These formulations are sentence-level methods that make individual predictions for each sentence. Recently, \citet{zhong2020extractive} observed that a summary consisting of
sentences with the highest scores is not necessarily the best. As a result, summary-level formulation like text matching~\cite{zhong2020extractive,an2023colo} and reinforcement learning~\cite{narayan2018ranking,bae2019summary} are proposed. Our proposed framework \model \ is also a novel summary-level extractive system with generation augmentation. Instead of sequentially labeling sentences, \model \ directly generates the desired summary sentence representations with diffusion models and extracts sentences by representation matching.

\subsection{Diffusion Models on Text}
 Continuous diffusion models are first introduced in \cite{sohl2015deep} and have achieved great success in continuous domain generations like image, video, and audio \cite{kong2020diffwave, yang_diffusion_2022, rombach2022high, ho2022video}. Nevertheless, few works have applied continuous diffusion model to text data due to its inherently discrete nature. Among the initial attempts, Diffusion-LM~\cite{li_diffusion-lm_2022} first adapts continuous diffusion models for text by adding an embedding step and a rounding step, and designing a training objective to learn the embedding. DiffuSeq~\cite{gong_diffuseq_2022} proposes a diffusion model designed for sequence-to-sequence (seq2seq) text generation tasks by adding partial noise during the forward process and conditional denoising during the reverse process. CDCD~\cite{dieleman2022continuous} is
proposed for text modeling and machine translation based on variance-exploding stochastic differential equations (SDEs) on token embeddings. SeqDiffuSeq~\cite{yuan2022seqdiffuseq} also proposes an encoder-decoder diffusion model architecture for conditional generation by combining self-conditioning and adaptive noise schedule technique. However, these works only focus on generating token-level embeddings for short text generation (less than 128 tokens). In order to adapt diffusion models to longer sequences like summaries, our \model \ directly generates summary sentence embeddings with a partial denoising framework. In addition, \model \ jointly optimizes the diffusion model with a contrastive sentence encoding module instead of using a static embedding matrix.

\section{Preliminary}
\subsection{Continuous Diffusion Models}
\label{sec:diffusion}
The continuous diffusion model \cite{ho2020denoising} is a probabilistic model containing two Markov chains: the forward and the backward process. 

\noindent
\textbf{Forward Process} Given a data point sampled from a real-world data distribution $\mathbf{x}_0 \sim q(x)$, the forward process gradually corrupts $\mathbf{x}_0$ into a standard Gaussian distribution prior $\mathbf{x}_{T} \sim \mathcal{N}(\mathbf{0}, \mathbf{I})$. Each step of the forward process gradually interpolates Gaussian noise to the sample, represented as:
\begin{equation}
\label{eq:diffusion}
   q(\mathbf{x}_{t+1} | \mathbf{x}_{t}) = \mathcal{N}\left(\mathbf{x}_{t+1} ; \sqrt{1-\beta_{t}} \mathbf{x}_{t}, \beta_{t} \mathbf{I}\right),
\end{equation}
where $\beta_{t} \in (0, 1)$ adjusts the scale of the variance. 

\noindent
\textbf{Reverse Process}~The reverse process starts from $\mathbf{x}_{T} \sim \mathcal{N}(0,I)$ and learns a parametric distribution $p_{\theta} \left(\mathbf{x}_{t-1} | \mathbf{x}_{t}\right)$ to invert the diffusion process of  Eq.~\ref{eq:diffusion} gradually. Each step of the reverse process is defined as:
\begin{equation}
    p_{\theta} \left(\mathbf{x}_{t-1} | \mathbf{x}_{t}\right) = \mathcal{N}\left(\mathbf{x}_{t-1} ; \mu_\theta\left(\mathbf{x}_t, t\right), \sigma^2_\theta\left( t\right)\mathbf{I}\right),
\end{equation}
where $\mu_\theta\left(\mathbf{x}_t, t\right)$ and $\sigma^2_\theta(t)$ are
learnable means and variances predicted by neural networks.

While there exists a tractable variational lower-bound (VLB) on $\log p_\theta\left(\mathbf{x}_0\right)$, \citet{ho2020denoising} simplifies the loss function of continuous diffusion to: 
\begin{equation}
\label{eq:simple}
\begin{aligned} 
\mathcal{L}_{\text{simple}} &= \sum_{t=1}^T\left\|\mathbf{x}_0-\tilde{f}_\theta\left(\mathbf{x}_t, t\right)\right\|^2,
\end{aligned}
\end{equation}
where $\tilde{f}_\theta\left(\mathbf{x}_t, t\right)$ is the reconstructed  $\mathbf{x}_0$ at step $t$.



\subsection{Problem Formulation}
\begin{figure}
    \centering
    \includegraphics[width=0.38\textwidth]{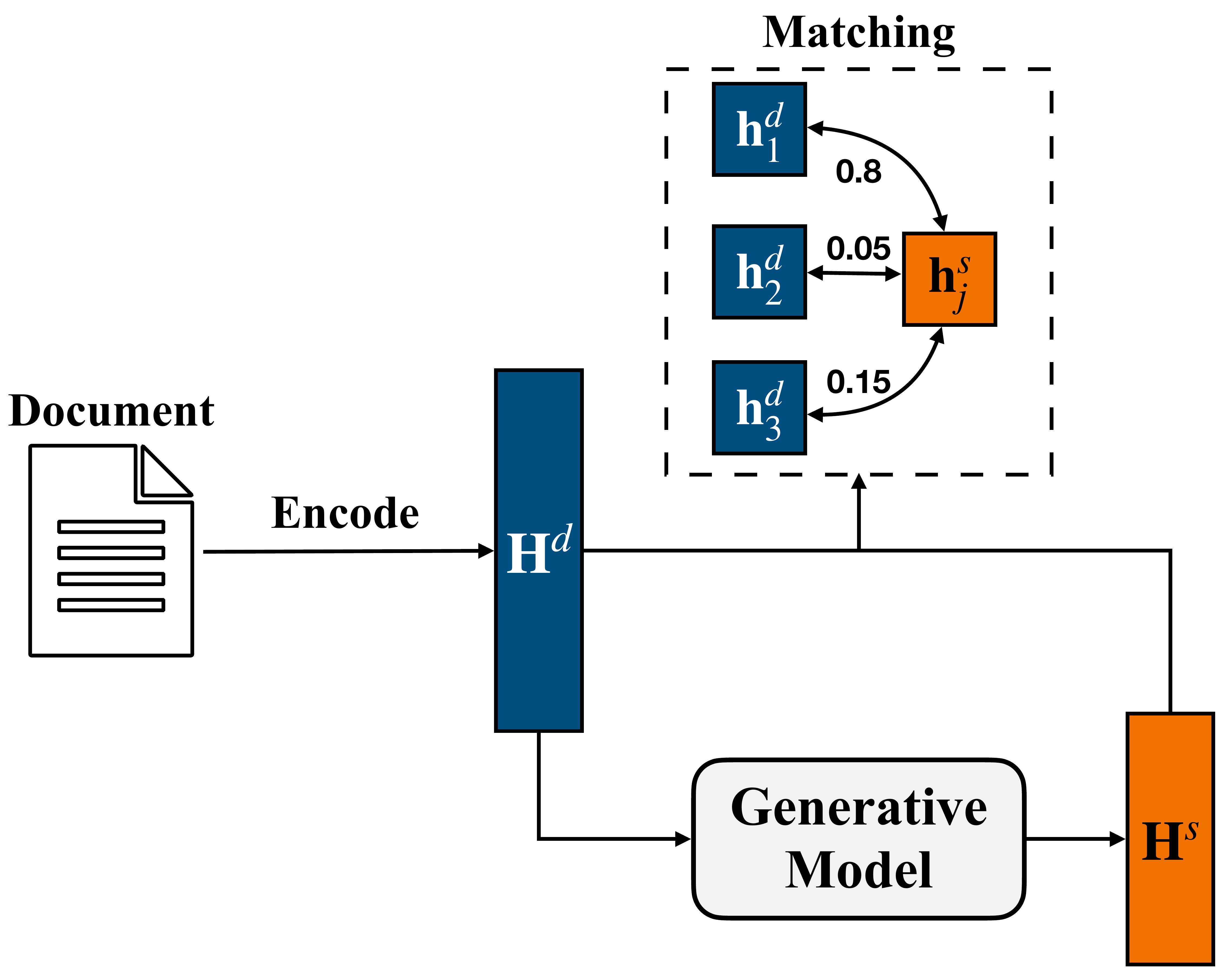}
    \caption{The proposed generation-enhanced extractive summarization framework. The model first conditionally generates desired summary embeddings and then extracts sentences based on representation matching.}
    \label{background}
\end{figure}

\begin{figure*}[t]
    \centering
    \vspace{-10pt}\includegraphics[width=0.9\textwidth]{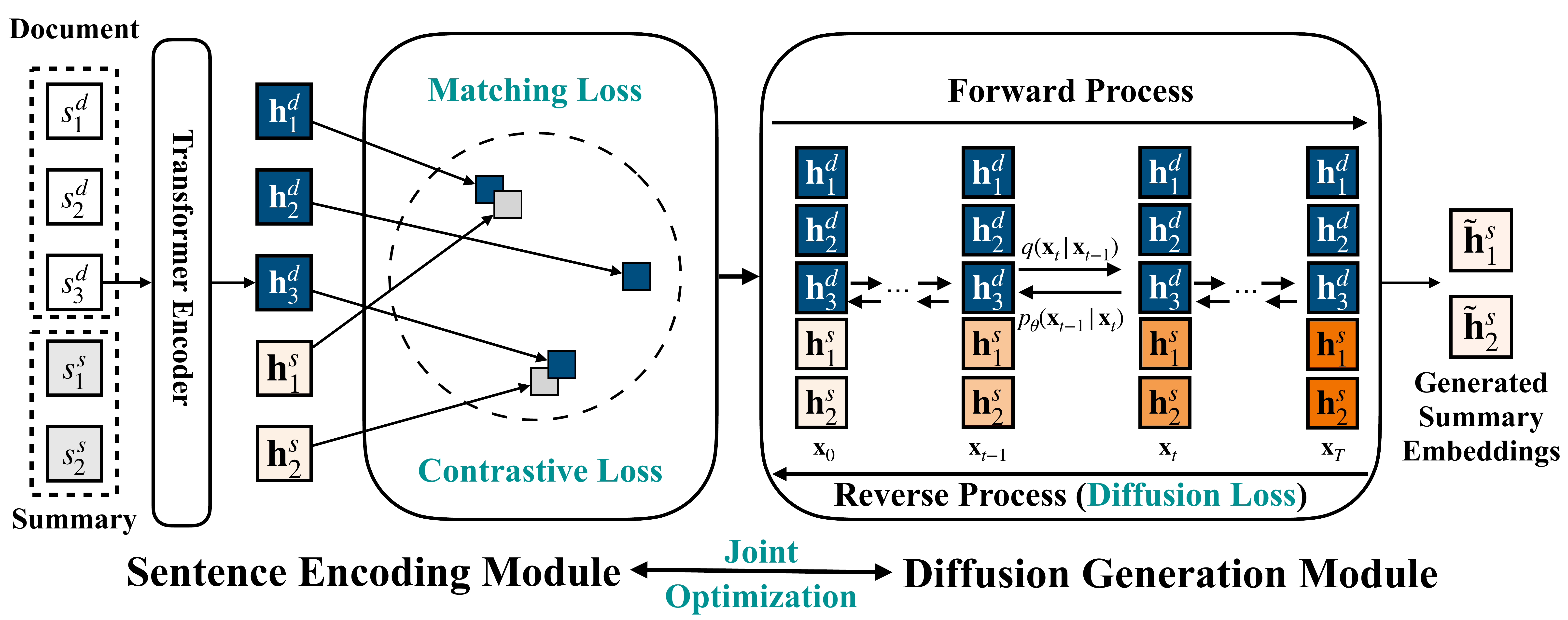}
    \caption{The overall architecture of {\model}. The input document is passed to the sentence encoding module and the diffusion generation module. {\model} will generate the desired summary sentence representations for inference.}
    \label{model_arch}
\end{figure*}
Given a document with $n$ sentences as $D=\{s_1^d, s_2^d, ..., s_n^d\}$, extractive summarization system aims to form a $m (m \ll n)$ sentences summary $S=\{s_1^s, s_2^s, ..., s_m^s \}$ by directly extracting sentences from the source document. Most existing work formulates it as sequence labeling and gives each sentence a $\{0,1\}$ label, where label $1$ indicates that the sentence will be included in summary $S$. Since extractive ground-truth labels (ORACLE) are not available for human-written gold summary, it is common to use a greedy algorithm to generate an ORACLE
consisting of multiple sentences which maximize
the ROUGE-2 score against the gold summary following \cite{nallapati2017summarunner}.

In contrast, we propose a summary-level framework with generative model augmentation as shown in Figure~\ref{background}. Formally, we train a diffusion model with the \textit{reverse process} $p_\theta({\tilde{\mathbf{H}}^s_{t-1}}|{\tilde{\mathbf{H}}^s_{t}}, \mathbf{H}^d)$ to directly generate the desired summary sentence representations $\tilde{\mathbf{H}}^s_{t-1}  = [{\tilde{\mathbf{h}}_1^s}, {\tilde{\mathbf{h}}_2^s}, ..., {\tilde{\mathbf{h}}_m^s}] \in \mathbb{R}^{m \times h}$, where ${\tilde{\mathbf{h}}_j^s}$ is the vector representing the j-th summary sentence at diffusion step $t-1$. The model then extracts summary sentences based on the matching between the generated summary sentence representations after $T$ reverse steps $\tilde{\mathbf{H}}^s_0 = [{\tilde{\mathbf{h}}_1^s}, {\tilde{\mathbf{h}}_2^s}, ..., {\tilde{\mathbf{h}}_m^s}]$ and the document sentence embeddings $\mathbf{H}^d = [\mathbf{h}_1^d, \mathbf{h}_2^d, ..., \mathbf{h}_n^d]$. The matching score for the j-th sentence in the output $s_j^s$ with the document is defined as:
\begin{equation}
\label{eq:pred}
    \tilde{\mathbf{y}}_j = \operatorname{softmax}({\tilde{\mathbf{h}}}^s_j \cdot \mathbf{H}^{d^{T}}).
\end{equation}
Here we use dot product as similarity measurement and then extract the sentence with the highest matching score for each generated summary sentence.

Our framework operates on the summary level by generating all summary sentence representations simultaneously and adopts continuous diffusion models here for sentence embedding generation.




\section{Method}


In this section, we introduce the detailed design of \model. \model \ consists of two major modules: a sentence encoding module and a diffusion module, which will be introduced in Section~\ref{sen_enco} and Section~\ref{diffu_procs}, respectively. After that, we explain how we optimize our model and conduct inference in Section~\ref{inference}. The overall model architecture of \model \ is also illustarted in Figure \ref{model_arch}.

\subsection{Sentence Encoding Module}
\label{sen_enco}
In order to generate desired summary sentence embeddings, we first build a contrastive sentence encoding module to transfer discrete text inputs $D = \{s_1^d, s_2^d, ..., s_n^d\}$ into continuous vector representations $\textbf{H}^d = [\textbf{h}_1^d, \textbf{h}_2^d, ..., \textbf{h}_n^d] \in \mathbb{R}^{n \times h}$, where $h$ is the dimension of the encoded sentence representations.

Specifically, we first obtain the initial representations of sentences $\mathbf{E}^d = [\mathbf{e}_1^d, \mathbf{e}_2^d, ..., \mathbf{e}_n^d ]$ with Sentence-BERT \cite{reimers2019sentence}. Note that the Sentence-BERT is
only used for initial sentence embedding, but is not
updated during training. The initial representations are then fed into a stacked transformer layer followed by a projection layer to obtain contextualized sentence representations $\textbf{h}_i^d$:
\begin{align}
    \mathbf{h}_i^d &= \operatorname{MLP}(\operatorname{Transformer}(\mathbf{e}_i^d)).
\end{align}
The same encoding process is applied to the summary sentences $S=\{s_1^s, s_2^s, ..., s_m^s \}$ to obtain encoded summary sentence representations $\textbf{H}^s = [\textbf{h}_1^s, \textbf{h}_2^s, ..., \textbf{h}_m^s] \in \mathbb{R}^{m \times h}$. The encoded document sentence representations $\textbf{H}^d$ and summary sentence representations $\textbf{H}^s$ are then concatenated as $\textbf{H}^{in} = \textbf{H}^d \| \textbf{H}^s \in \mathbb{R}^{(n+m) \times h}$ and will be passed to the diffusion generation module.

To ensure the sentence encoding module produces accurate and distinguishable representations, we introduce a matching loss $\mathcal{L}_{\text{match}}$ and a multi-class supervised contrastive loss $\mathcal{L}_{\text{contra}}$ to optimize the module, which are defined as follows.:

\noindent
\textbf{Matching Loss} We first introduce a matching loss to ensure an accurate matching between the encoded document and summary sentence representations. Formally, for the j-th encoded summary sentence representation $\textbf{h}^s_j$, we generate its encoding matching scores $\mathbf{\hat{y}}_j$ by computing the dot product with document representations followed by a softmax function:
\begin{equation}
    \begin{split}
        \hat{\mathbf{y}}_j = \operatorname{softmax}(\mathbf{h}_j^s \cdot \mathbf{H}^{d ^\mathrm{T}}). \\ 
    \end{split}
    \label{equ:pred}
\end{equation}
Then we have the encoding matching loss $\mathcal{L}_{\text{match}}$ as the cross-entropy between our encoding matching score $\hat{\mathbf{y}}_j$ and the ground truth extractive summarization label (ORACLE) $\mathbf{y}_j$: 
\begin{equation}
    \begin{aligned}
    \mathcal{L}_{\text{match}} &= \sum_{j=1}^{m}\operatorname{CrossEntropy}\left(\mathbf{y}_j, \mathbf{\hat{y}}_j \right). 
    \end{aligned}
\end{equation}

\noindent
\textbf{Contrastive Loss} The sentence encoding module also needs to ensure the encoded summary sentence embeddings $[\textbf{h}_1^s, \textbf{h}_2^s, ..., \textbf{h}_m^s]$ are diverse and distinguishable. Thus, we introduce the multi-class supervised contrastive loss \cite{khosla2020supervised} to push the summary sentence representation closer to its corresponding document sentence representation while keeping it away from other sentence embeddings. 

Given the sentence contextual representations $\mathbf{H}^{in}=[\mathbf{h}_1, \mathbf{h}_2, ..., \mathbf{h}_{n+m}] \in \mathbb{R}^{(n+m) \times h}$, the contrastive label $\mathbf{y}^c$ is defined as:
\begin{equation}
    y_p^{c} =  \left\{\begin{array}{l}
q, \text { if } p \leq n \text{ and } s_p^d=s_q^s \\
q, \text{ if }  p = n+q \\
0, \text { otherwise }
\end{array}\right. ,
\end{equation}
where $q \in \{1, 2, \cdots, m\}$ and $y_p^c$ is the p-th element of $\mathbf{y}^c$. The contrastive loss $\mathcal{L}_{\text{contra}}$ is defined as:

\begin{equation}
\small
\begin{aligned}
\mathcal{L}_{\text {contra }} & = \frac{-1}{2 N_{y^c_p}-1} \sum_{p=1}^{n + m}  \mathcal{L}^p_{\text {contra }}, \\
\mathcal{L}^p_{\text{contra}} & = \sum_{q=1;q \neq p; \atop y_q^c = y_p^c }^{n+m} \log \frac{\exp \left(\textbf{h}_p \cdot \textbf{h}_q^T / \tau\right)}{\sum_{k=1; p \neq k}^{n+m} \exp \left(\textbf{h}_p \cdot \textbf{h}_k^T / \tau\right)},
\end{aligned}
\end{equation}
where $N_{y_p^c}$ is the total number of sentences in the document that have the same label $y_p^c$ ($N_{y_p^c}=2$ in our case) and $\tau$ is a temperature hyperparameter.

The overall optimizing objective for the sentence encoding module $\mathcal{L}_{\text{se}}$ is defined as:
\begin{equation}
    \mathcal{L}_{\text{se}} = \mathcal{L}_{\text{match}} + \gamma\mathcal{L}_{\text{contra}}, 
\end{equation}
where $\gamma$ is a rescale factor that adjusts the diversity of the sentence representations. 


\subsection{Diffusion Generation Module}
\label{diffu_procs}
After obtaining the input encoding $\mathbf{H}^{in} = \mathbf{H}^d \| \mathbf{H}^s$, we adopt the continuous diffusion model to generate desired summary sentence embeddings conditionally. As described in Section~\ref{sec:diffusion}, our diffusion generation module adds Gaussian noise gradually through the forward process and fits a stacked Transformer to invert the diffusion in the reverse process.   

 We first perform one-step Markov transition $q(\mathbf{x}_0 | \mathbf{H}^{in}) = \mathcal{N} \left(\mathbf{H}^{in}, \beta_0 \mathbf{I}\right)$ for the starting state $\mathbf{x}_0 = \mathbf{x}_0^d\|\mathbf{x}_0^s$. Note that the initial Markov transition is applied to both document and summary sentence embeddings.

We then start the forward process by gradually injecting partial noise to summary embeddings $\mathbf{x}^s$ and leaving document embeddings unchanged $\mathbf{x}^d$ similar to \cite{gong_diffuseq_2022}. This enables the diffusion model to generate conditionally on the source document. At step $t$ of the forward process~$q\left( \mathbf{x}_{t}^s | \mathbf{x}_{t-1} \right)$, the noised representations is $\mathbf{x}_t$:
\begin{equation}
    \mathbf{x}_t =  \mathbf{x}_0^d ||  \mathcal{N}\left(\mathbf{x}_{t}^s ; \sqrt{1-\beta_t} \mathbf{x}_{t-1}^s, \beta_t \mathbf{I}\right),
\end{equation}
where 
$t\in\{1, 2, \cdots, T\}$ for a total of $T$ diffusion steps and $\|$ represents concatenation.  

Once the partially noised representations are acquired, we conduct the backward process to remove the noise of summary representations given the condition of the sentence representations of the previous step:
\begin{align}
    p_\theta  \left( \mathbf{x}_{t-1}^s | \mathbf{x}_{t} \right) =
    \mathcal{N}\left(\mathbf{x}_{t-1}^s ; \mu_\theta\left(\mathbf{x}_t, t\right), \sigma_\theta^2\left(t\right)\mathbf{I}\right),
\end{align}
where $\mu_\theta\left(\cdot \right)$ and $\sigma_\theta^2\left(\cdot \right)$ are parameterized models (stacked Transformer in our case) to predict the mean and standard variation at diffusion step $t-1$. The final output of the diffusion module is the generated summary sentence representations after $T$ reverse steps $\tilde{\mathbf{H}}^s_0 = [{\tilde{\mathbf{h}}_1^s}, {\tilde{\mathbf{h}}_2^s}, ..., {\tilde{\mathbf{h}}_m^s}]$. We optimize the diffusion generation module with diffusion loss $\mathcal{L}_{\text{diffusion}}$ defined as:
\begin{equation}
\label{eq:diff}
\begin{aligned} 
\mathcal{L}_{\text{diffusion}} &= \sum_{t=2}^T\left\|\mathbf{x}_0-\tilde{f}_\theta\left(\mathbf{x}_t, t\right)\right\|^2 +\\
& \left\|\mathbf{H}^{in}-\tilde{f}_\theta\left(\mathbf{x}_1, 1\right)\right\|^2 +\mathcal{R}\left(\mathbf{x}_0\right),
\end{aligned}
\end{equation}
where $\tilde{f}_\theta\left(\mathbf{x}_t, t\right)$ is the reconstructed $\mathbf{x}_0$ at step $t$ and $\mathcal{R}\left(\mathbf{x}_0\right)$ is a L-2 regularization term.

\subsection{Optimization and Inference}
\label{inference}

We jointly optimize the sentence encoding module and the diffusion generation module in an end-to-end manner. The overall training loss of {\model} can be represented as:
\begin{equation}
    \mathcal{L} = \mathcal{L}_{\text{se}}+\eta\mathcal{L}_{\text{diffusion}}
\end{equation}
where $\eta$ is a balancing factor of sentence encoding module loss $\mathcal{L}_{\text{se}}$ and diffusion generation module loss $\mathcal{L}_{\text{diffusion}}$.

For inference, \model \ first obtains encoded document representations $\mathbf{H}^d$, followed by a one-step Markov transition $q(\mathbf{x}^d_0 | \mathbf{H}^d)$. Then we random sample $m$ Gaussian noise embeddings as initial summary sentence representations $\mathbf{x}^s_{T}\in \mathbb{R}^{m \times h}$ and concatenate it with document representations to get the input $\mathbf{x}_T^{in} = \mathbf{x}_0^d || \mathbf{x}^s_{T}$ for diffusion step $T$. Then \model \ applies the learned reverse process (generation process) to remove the Gaussian noise iteratively and get the output summary sentence representations $\tilde{\mathbf{H}}^s_0 = [{\tilde{\mathbf{h}}_1^s}, {\tilde{\mathbf{h}}_2^s},..., {\tilde{\mathbf{h}}_m^s}]$.

\model \ then calculates the matching between the generated summary representation $\tilde{\mathbf{h}}_i^s$ and the document representation $\mathbf{H}^d$ to obtain prediction label $\mathbf{\tilde{y}}_i^{pred}$ as in Eq.~\ref{eq:pred}.
We extract the sentence with the highest score for each generated summary sentence representation and form the summary.

\section{Experiment}
\subsection{Experimental Setup}
\begin{table}[t]
\centering
\small
\scalebox{1}{
\begin{tabular}{c | c  c  c c }
    \hline \textbf{Dataset} & \textbf{Domain} & \begin{tabular}{@{}c@{}}\textbf{Doc}  \\ \textbf{\#words}\end{tabular} & \begin{tabular}{@{}c@{}}\textbf{Sum} \\\textbf{\#words}\end{tabular} & \textbf{\#Ext} \Tstrut\Bstrut\\
    \hline
    CNN/DM & News & 766.1 & 58.2 & 3 \Tstrut \\ 
    XSum & News & 430.2 & 23.3 & 2 \Bstrut \\ 
    PubMed & Paper & 444 & 209.5 & 6  \Bstrut \\
    \hline
\end{tabular}
}
\caption{Statistics of the experimental datasets. Doc \# words and Sum \# words refer to the average word number in the source document and summary. \# Ext refers to the number of sentences to extract.}
\label{data_stat}
\end{table}
\noindent
\textbf{Datasets} We conduct experiments on three benchmark summarization datasets: \textit{CNN/DailyMail}, \textit{XSum}, and~\textit{PubMed}. {CNN/DailyMail}~\cite{hermann2015teaching} is the most widely-adopted summarization dataset that contains news articles and corresponding human-written news highlights as summaries. We use the non-anonymized version in this work and follow the common training, validation, and testing splits (287,084/13,367/11,489). {XSum}~\cite{narayan2018don} is a one-sentence
news summarization dataset with all summaries professionally written by the original authors of the documents. We follow the common training, validation, and testing splits (204,045/11,332/11,334). {PubMed}~\cite{cohan2018discourse} is a scientific paper summarization dataset of long documents. We follow the setting in~\cite{zhong2020extractive} and use the introduction section as the article and the abstract section as the summary. The training/validation/testing split is (83,233/4,946/5,025). The detailed statistics of each dataset are shown in Table~\ref{data_stat}.\\

\noindent
\textbf{Baselines}
We compare \model \ with strong sentence-level baseline methods:
the vanilla Transformer~\cite{vaswani2017attention}, hierarchical encoder model HIBERT~\cite{zhang2019hibert}, PNBERT~\cite{zhong2019searching} that combines LSTM-Pointer with pre-trained BERT, BERT-based extractive model BERTSum~\cite{liu2019text}, and BERTEXT~\cite{bae2019summary} that augments BERT with reinforcement learning,.

We also compare \model \ with state-of-the-art summary-level approaches:
contrastive Learning based re-ranking framework COLO~\cite{an2023colo} and summary-level two-stage text matching framework MATCHSUM~\cite{zhong2020extractive}. 

\subsection{Implementation Details}
We use Sentence-BERT \cite{reimers2019sentence} checkpoint \textit{all-mpnet-base-v2} 
for initial sentence representations. The dimension of the sentence representations $h$ is set to $128$. We use an 8-layer Transformer with 12 attention heads in our sentence encoding module and a 12-layer Transformer with 12 attention heads in the diffusion generation module. The hidden size of the model is set to $768$, and temperature $\tau$ is set to $0.07$. The scaling factors $\gamma$ and $\eta$ are set to 0.001 and 100, where $\gamma$ is searched in the range of $[0.0001, 1]$ and $\eta$ is searched within the range of $[10, 1000]$. We set the diffusion steps $T$ to 500. Effects of hyperparameter $T$ and $h$ are discussed in section~\ref{sec:hyper}.

{\model} has a total of 13 million parameters and is optimized with AdamW optimizer \cite{loshchilov2017decoupled} with a learning rate of $1e^{-5}$ and a dropout rate of $0.1$. We train the model for $10$ epochs and validate the performance by the average of ROUGE-1 and ROUGE-2 F-1 scores on the validation set. 

Following the standard setting, we evaluate model performance with ROUGE\footnote{ROUGE: https://pypi.org/project/rouge-score/} F-1 scores~\cite{lin2003automatic}. Specifically, ROUGE-1/2 scores measure summary informativeness, and the ROUGE-L score measures summary fluency. Single-run results are presented in the following sections with the default random seed of 101.

\begin{table}[t]
\centering
\scalebox{0.83}{
\begin{tabular}{l  c  c  c }
    \hline \textbf{Model} & \textbf{R-1} & \textbf{R-2} & \textbf{R-L} \Tstrut\Bstrut\\
    \hline
    LEAD & 40.43 & 17.62 & 36.67 \Tstrut\Bstrut \\ 
    ORACLE & 52.59 & 31.23 & 48.87  \Tstrut\Bstrut \\ 
    \hline
\rowcolor{gray!10}
    \multicolumn{4}{c}{\textbf{\textsl{One-stage Systems}}} \\
    \hline
    Transformer (\citeyear{vaswani2017attention}) & 40.90 & 18.02 & 37.17 \Tstrut \Bstrut\\
    $\text{HIBERT}^*$ (\citeyear{zhang2019hibert}) & 42.37 & 19.95 & 38.83 \Tstrut \Bstrut \\
    $\text{PNBERT}^*$ (\citeyear{zhong2019searching}) & 42.69 & 19.60 & 38.85 \Tstrut \Bstrut \\
    $\text{BERTEXT}^*$ (\citeyear{bae2019summary}) & 42.76 & 19.87 & 39.11  \Tstrut \Bstrut \\
    $\text{BERTSum}^*$ (\citeyear{liu2019text}) & 43.85 & 20.34 & 39.90 \Tstrut \Bstrut \\
    $\text{COLO}_{\text{Ext}}^*$ (\citeyear{an2023colo}) & 44.58 & 21.25 & 40.65 \Tstrut \Bstrut \\
    {\model} (ours) & \textbf{44.62} & \textbf{22.51} & \textbf{40.34} \Tstrut \Bstrut \\
    \hline
    
\rowcolor{gray!10}
\multicolumn{4}{c}{\textbf{\textsl{Two-stage Systems}}} \\

\hline$\text{MATCHSUM}^*\text{(BERT)}$ 

(\citeyear{zhong2020extractive}) & 44.22 & 20.62 & 40.38 \Tstrut \Bstrut \\
    $\text{MATCHSUM}^*\text{(Roberta)}$ & 44.41 & 20.86 & 40.55 \Tstrut \Bstrut \\
    {\model} (ours)& \textbf{44.83} & \textbf{22.56} & \textbf{40.56}  \Tstrut\Bstrut \\
    \hline
\end{tabular}
}

\caption{Experimental results on CNN/DailyMail dataset. Models using pre-trained language models are marked with*.}
\label{main_exp_cnn}
\end{table}
\subsection{Experiment Results}
\begin{table}[t]
\centering
\scalebox{0.8}{

\begin{tabular}{l | c  c  c | c  c  c  }
    \hline \multirow{2}*{\textbf{Model}} & \multicolumn{3}{c|}{\textbf{{PubMed}}}  & \multicolumn{3}{c}{\textbf{{XSum}}} \Tstrut\Bstrut\\
    ~ & \textbf{{R-1}} & \textbf{{R-2}} & \textbf{{R-L}}  & \textbf{{R-1}} & \textbf{{R-2}} & \textbf{{R-L}}   \Tstrut\Bstrut\\
    \hline 
        ORACLE & 45.12 & 20.33 & 40.19  & 25.62 & 7.62 & 18.72 \Tstrut\\
        LEAD & 37.58 & 12.22 & 33.44  & 14.40 & 1.46 & 10.59  \Bstrut\\
    \hline 
    BERTSUM & 41.05 & 14.88 & 36.57  & 22.86 & 4.48 & 17.16  \Tstrut \\
    MatchSUM & 41.21 & 14.91 & 36.75  & \textbf{24.86} & 4.66 & \textbf{18.41} \\
    {\model} & \textbf{41.40}  & \textbf{15.55} & \textbf{37.48} & 24.00 & \textbf{5.44} & 18.01 \Bstrut\\
    \hline
\end{tabular}

}

\caption{Experimental Results on PubMed and XSum datasets.}

\label{exp_result}
\end{table}

\noindent
\textbf{Results on CNN/DailyMail} Experimental results on CNN/DailyMail dataset are shown in Table \ref{main_exp_cnn}. The first block in the table contains the extractive ground truth ORACLE (upper bound) and LEAD that selects the first few sentences as a summary. The second block includes recent strong one-stage extractive baseline methods and our proposed model \model. The third section includes two-stage baseline methods that pre-select salient sentences. We follow the same setting and show the results of {\model} with the same pre-selection for a fair comparison.

According to the results, \model \ achieves \textit{new state-of-the-art} performance under both one-stage and two-stage settings, especially a large raise in the ROUGE-2 score. The supreme performance of {\model} demonstrates the efficacy of our generation-augmented framework and the great potential to apply diffusion models in text representation generation. It is worth noting that most baseline methods contain pre-trained language model components, but our proposed framework \model \ trains Transformers from scratch and contains no pre-trained knowledge. We believe \model \ would achieve even better performance if combining pre-trained knowledge, and leave it to future work. We also notice that summary-level methods generally outperform sentence-level methods, proving the need to fill the inherent gap. 

\noindent
\textbf{Results on XSum and PubMed}
We also evaluate \model \ on PubMed and XSum datasets, representing datasets of different domains and summary lengths as shown in Table \ref{exp_result}.

For data with longer summaries like PubMed, \model \ shows highly strong performance and outperforms state-of-the-art baselines. The strong performance proves that our model can tackle longer input contexts and complex generations. Our summary-level setting also benefits data with longer summaries by considering summary sentence dependencies. 

For data with shorter summaries like XSum, \model \ also achieves comparable performance to SOTA approaches, with a significantly higher ROUGE-2 score. Short-summary data tend to be simpler for matching-based methods like MatchSum since the candidate pool is much smaller.

Overall, \model \ achieves a comparable or even better performance compared to pre-trained language
model-based baseline methods. The results demonstrate the effectiveness of {\model} on summarization data with different lengths.

\section{Analysis}
\subsection{Ablation Study}
\label{sec:ablation}
\begin{table}[t]
\centering
\scalebox{1}{
\begin{tabular}{l | c  c  c  }
    \hline \textbf{Model} & \textbf{R-1} & \textbf{R-2} & \textbf{R-L} \Tstrut\Bstrut\\
    \hline
   {\model} & \textbf{44.83} & \textbf{22.56} & \textbf{40.56} \Tstrut\Bstrut \\
    w/o Sentence-BERT & 43.53 & 21.63 & 40.23 \Bstrut \\
        w/o ORACLE & 39.19&  17.12& 34.38 \Bstrut \\
    w/o Contrastive Loss & 44.57 & 22.35 & 40.34  \Bstrut\\
    \hline
\end{tabular}
}

\caption{Ablation study results on CNN/DailyMail dataset.}
\label{ablation}
\end{table}
To understand the strong performance of \model, we perform an ablation study by removing model components of the sentence encoding module and show the results in Table~\ref{ablation}. The second row shows that performance drops when replacing the initial sentence representation from Sentence-BERT to BERT-base encoder\cite{devlin2018bert}. The performance drop indicates sentence-level information is necessary for the success of {\model}. The third row shows that replacing ORACLE with abstractive reference summaries degrades performance. As for the sentence encoding loss, both the matching loss and contrastive loss benefit the overall model performance according to rows 4 and 5. The matching loss is critical to the model, and the performance drops dramatically by more than 40\% without it. The results prove the importance of jointly training a sentence encoder that produces accurate and diverse sentence representations with the generation module..

\subsection{Hyperparameter Sensitivity}
\label{sec:hyper}
\begin{table}[!htbp]
\centering
\small
\scalebox{1}{
\begin{tabular}{l  c  c  c  }
    \hline \textbf{Model} & \textbf{R-1} & \textbf{R-2} & \textbf{R-L} \Tstrut\Bstrut\\
    \hline
    \model($T$=500, $h$=128) & \textbf{44.83} & \textbf{22.56} & \textbf{40.58} \Tstrut\Bstrut\\
    \hline
    \model($T$=500, $h$=64) & 43.36 & 21.27 & 39.89 \Tstrut\Bstrut\\
    \model($T$=500, $h$=256) & 44.53 & 22.49 & 40.27 \Bstrut\\
    \hline
    \model($T$=50, $h$=128) & 42.60 & 19.71 & 38.96 \Tstrut\Bstrut\\
    \model($T$=100, $h$=128) & 44.61 & 22.24 & 40.32 \Bstrut\\
    \model($T$=1000, $h$=128)   & 44.65 & 22.36 & 40.37 \Bstrut\\
    \model($T$=2000, $h$=128)  & 44.64 & 22.37 & 40.40 \Bstrut\\
    
    \hline
\end{tabular}
}

\caption{The performance of {\model} with different hyperparameter settings on CNN/DM dataset.}
\label{hyperparam}
\end{table}
\begin{table}[!htbp]
\centering
\small
\scalebox{0.83}{
\begin{tabular}{c | c  c  c }
    \hline \diagbox{\textbf{Train}}{\textbf{Test} ~\vspace{-5pt}} \Tstrut& \textbf{CNN/DM} & \textbf{XSum} & \textbf{PubMed} \Tstrut\Bstrut\\
    \hline
    \textbf{CNN/DM} & {44.83/22.56} &  21.35/3.85 & {39.83(-1.57)/13.25} \Tstrut \\ 
    \textbf{XSum} & 42.85/21.37 & {24.0/5.44} & 38.71(-2.69)/12.93 \Bstrut \\ 
    \hline
\end{tabular}
}
\caption{ROUGE-1 and ROUGE-2 results for cross-dataset evaluation.}
\label{adapt}
\end{table}
We also study the influence of our diffusion generation module's two important hyperparameters: diffusion steps $T$ and the sentence representations dimension $h$ in Table \ref{hyperparam}. The first row is our best model, and the second block shows the performance of \model \ with different sentence representation dimensions. The performance drops by a large margin when setting the dimension to $64$, indicating severe information loss when shrinking the sentence dimension too much. The performance also drops a little when the dimension is set to $256$, suggesting that a too-large dimension may bring in more noise. The third block shows the influence of diffusion steps, where we find that model performance first increases with more diffusion steps, then starts to decrease and oscillate if steps keep increasing. We argue that the noise injected in the forward pass cannot be fully removed if the steps are too small, and the model will introduce too much noise to recover if the steps are too big.

\subsection{Cross-dataset Evaluation}
We also notice that \model \ shows a strong cross-dataset adaptation ability. As shown in Table~\ref{adapt}, the model trained on the news domain (CNN/DM and XSum) achieves comparable performance (only 1.57 and 2.69 ROUGE-1 drops) when directly tested on the scientific paper domain. The cross-dataset results demonstrate the robustness of our generation-augmented framework and the potential to build a generalized extractive summarization system.
\begin{figure}
\vspace{-10pt}
    \centering
    \includegraphics[width=0.34\textwidth]{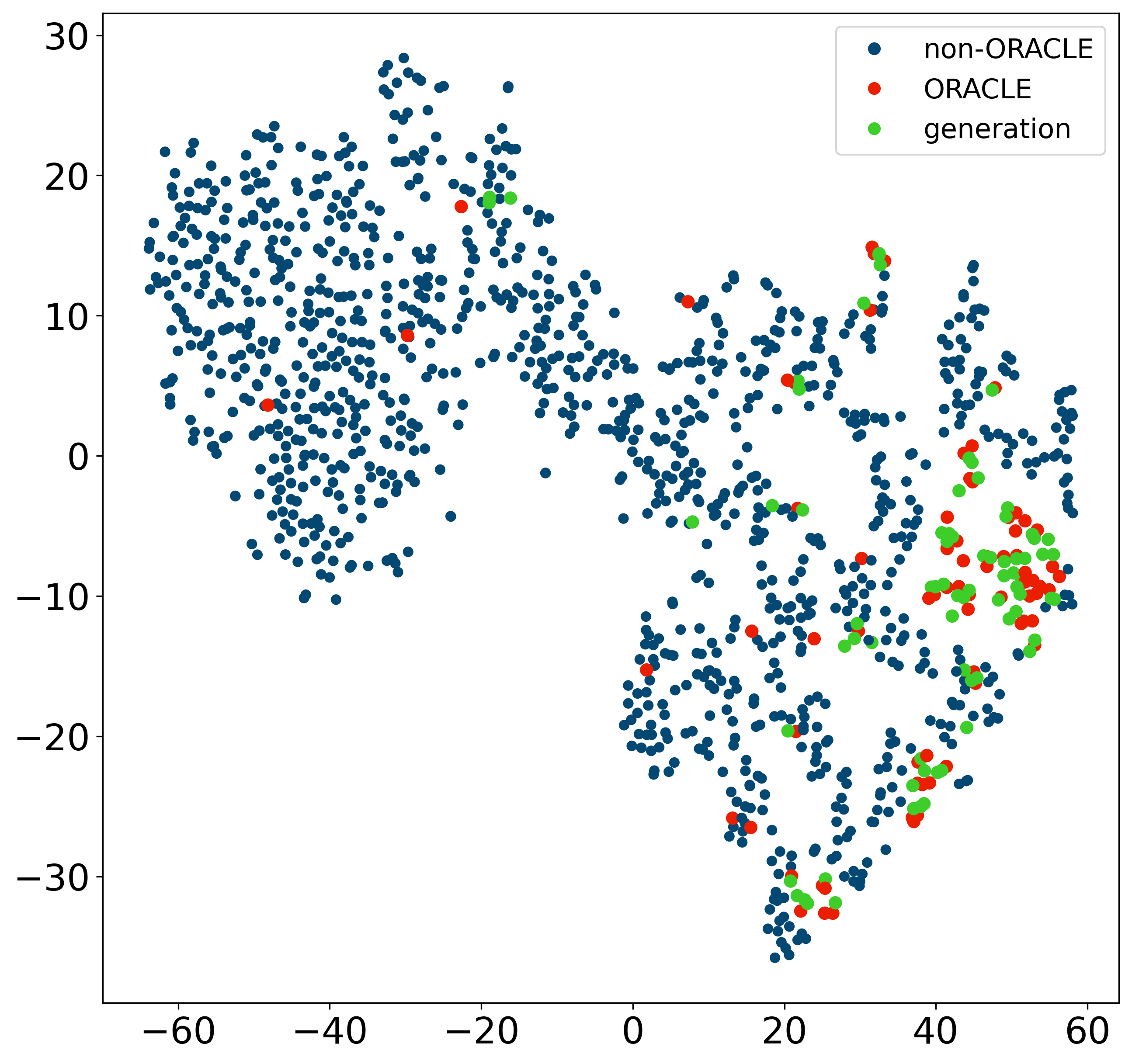}
    \caption{T-SNE visualization of sentence embeddings from 25 CNN/DM dataset documents.}
    \label{rep_vis}
\end{figure}
\subsection{Representation Analysis}
\label{sec:visual}
We also analyze the generated sentence representation quality. We apply T-SNE~\cite{JMLR:v9:vandermaaten08a} to reduce the sentence representation's dimension to 2 and show the encoded sentence representations as well as the generated summary sentence representations in Figure~\ref{rep_vis}. The blue dots in the figure represent non-summary sentences, and the red dots represent summary sentences (ORACLE) from our sentence encoding module. The green dots are summary sentence representations reconstructed by our diffusion generation module. We can find that most of the ORACLE sentences gather on the right. This finding proves that our contrastive encoder could distinguish ORACLE sentences from non-summary sentences. We also find that the sentence representations generated by the diffusion module (green) are very close to the original summary representations (red). The finding demonstrates that our diffusion generation module is powerful in reconstructing sentence representations from random Gaussian noise.


\section{Conclusions}

This paper proposes a new paradigm for extractive summarization with generation augmentation. Instead of sequentially labeling sentences, \model \ directly generates the desired summary sentence representations with diffusion models and extracts summary sentences based on representation matching. Experimental results on three benchmark datasets prove the effectiveness of \model. This work is the first attempt to adapt diffusion models for summarization. Future work could explore various ways of applying continuous diffusion models to both extractive and abstractive summarization.
\section*{Limitations}
Despite the strong performance of \model, its design still has the following limitations. First, \model \ is only designed for extractive summarization, and the diffusion generation module only generates sentence embeddings instead of token-level information. Thus, it is not applicable to the abstractive summarization setting. Moreover, \model \ is only tested on single document summarization datasets. How to adapt \model \ for multi-document and long document summarization scenarios need further investigation. In addition, our generative model involves multiple steps of noise injection and denoising, compared to discriminator-based extractive systems.
\section*{Acknowledgement}
This work is supported by NSF through grants IIS-1763365 and IIS-2106972. We also thank anonymous reviewers for their helpful feedback.

\bibliography{custom}
\bibliographystyle{acl_natbib}



\end{document}